\documentclass{article}

\PassOptionsToPackage{numbers, compress}{natbib}
 \usepackage[preprint]{neurips_2026}

\usepackage[utf8]{inputenc} 
\usepackage[T1]{fontenc}    
\usepackage{hyperref}       
\usepackage{url}            
\usepackage{booktabs}       
\usepackage{amsfonts}       
\usepackage{nicefrac}       
\usepackage{microtype}      
\usepackage{xcolor}         

\usepackage{amsmath}
\usepackage{amssymb}
\usepackage{url}
\usepackage{bbm}
\usepackage{multirow}
\usepackage{booktabs}
\usepackage{subfig}
\usepackage{blindtext}
\usepackage{wasysym}
\usepackage{pifont}
\usepackage{makecell}
\usepackage{graphicx}
\usepackage{enumitem}

\definecolor{darkgreen}{RGB}{0, 100, 0}
\definecolor{purple}{RGB}{128,0,128}
\definecolor{maroon}{RGB}{204,0,0}
\definecolor{navyblue}{RGB}{26,69,149}
\definecolor{Gray}{gray}{0.9}

\definecolor{green}{HTML}{4CBB17}

\newcommand{\policy}{\pi}
\newcommand{\user}{\mathcal{U}}
\newcommand{\env}{\varepsilon}
\newcommand{\goal}{g}
\newcommand{\outcomespace}{\Omega}

\newcommand{\envstatespace}{\mathcal{S}^{\env}}

\newcommand{\actionspace}{\mathcal{A}}
\newcommand{\obsspace}{\mathcal{O}}
\newcommand{\envobsspace}{\mathcal{O}^{\env}}
\newcommand{\userobsspace}{\mathcal{M}}
\newcommand{\envactionspace}{\mathcal{A}^{\env}}
\newcommand{\useractionspace}{\mathcal{A}^{\user}}
\newcommand{\outcome}{\omega}
\newcommand{\state}[1]{s_{#1}}
\newcommand{\envstate}[1]{s^{\env}_{#1}}

\newcommand{\obs}[1]{o_{#1}}
\newcommand{\action}[1]{a_{#1}}
\newcommand{\conv}[1]{c_{#1}}

\newcommand{\history}[1]{h_{#1}}
\newcommand{\pref}[1]{p_{#1}}
\newcommand{\prefset}{\mathcal{P}}
\newcommand{\respond}{\texttt{respond}}
\newcommand{\prefspace}{\mathbb{P}}
\newcommand{\reward}{r}
\newcommand{\completion}{\textsc{TaskCompletion}}
\newcommand{\outcomescore}{\textsc{OutcomeScore}}
\newcommand{\taskscore}{\textsc{TaskScore}}
\newcommand{\perfectoutcome}{\textsc{PerfectOutcome}}
\newcommand{\nulloutcome}{\varnothing}

\newcommand{\full}{\textcolor{blue}{\ding{51}}}
\newcommand{\partialencoded}{\textcolor{purple}{\LEFTcircle}}
\newcommand{\none}{\textcolor{red}{\ding{55}}}

\newcommand{\benchmarkname}{\textsc{AcCoRD}}
\newcommand{\benchmarknamefull}{\textbf{Ac}tive \textbf{Co}llaboration with \textbf{R}ealistic Preference \textbf{D}ynamics}

\title{\benchmarkname: Evaluating User-Agent Collaboration\\Under Realistic User Preference Dynamics}

\author{
    Tejas Srinivasan$^1$ \quad Shikib Mehri$^2$ \quad Nandita Naik$^2$ \\{\bf Anirban Das$^3$ \quad William M. Campbell$^3$ \quad Jesse Thomason$^4$} \\
  $^1$University of Southern California \quad
  $^2$Contextual AI \\
  $^3$Capital One \quad
  $^4$Georgia Tech University \\
  \texttt{tejas.srinivasan@usc.edu}
}

\begin{document}

\maketitle

\begin{abstract}
User preferences in user-agent collaboration are rarely static and fully-specified upfront: preferences are formed, revealed, adjusted, and relaxed during interaction. 
Existing benchmarks for evaluating user-agent collaboration focus almost exclusively on resolving underspecified preferences, thereby failing to capture the richer dynamics of real-world interaction. 
We introduce \benchmarkname, a user-agent collaboration benchmark requiring agents to handle diverse user preference dynamics in two domains: online shopping and travel planning. 
We evaluate five frontier LLMs under two prompting strategies: vanilla ReAct, and an uncertainty-guided variant that prompts models to identify and resolve ambiguity about user preferences. 
Our results reveal that frontier models can handle underspecification but struggle to satisfy preferences that emerge or evolve mid-interaction and require more sophisticated uncertainty modeling. 
Further, prompting alone fails to elicit the required uncertainty recognition. 
We release \benchmarkname\ as a resource for developing agents that can navigate the full complexity of real-world user preferences.
\end{abstract}

\section{Introduction}
\begin{figure}
    \centering
    \includegraphics[width=0.85\linewidth]{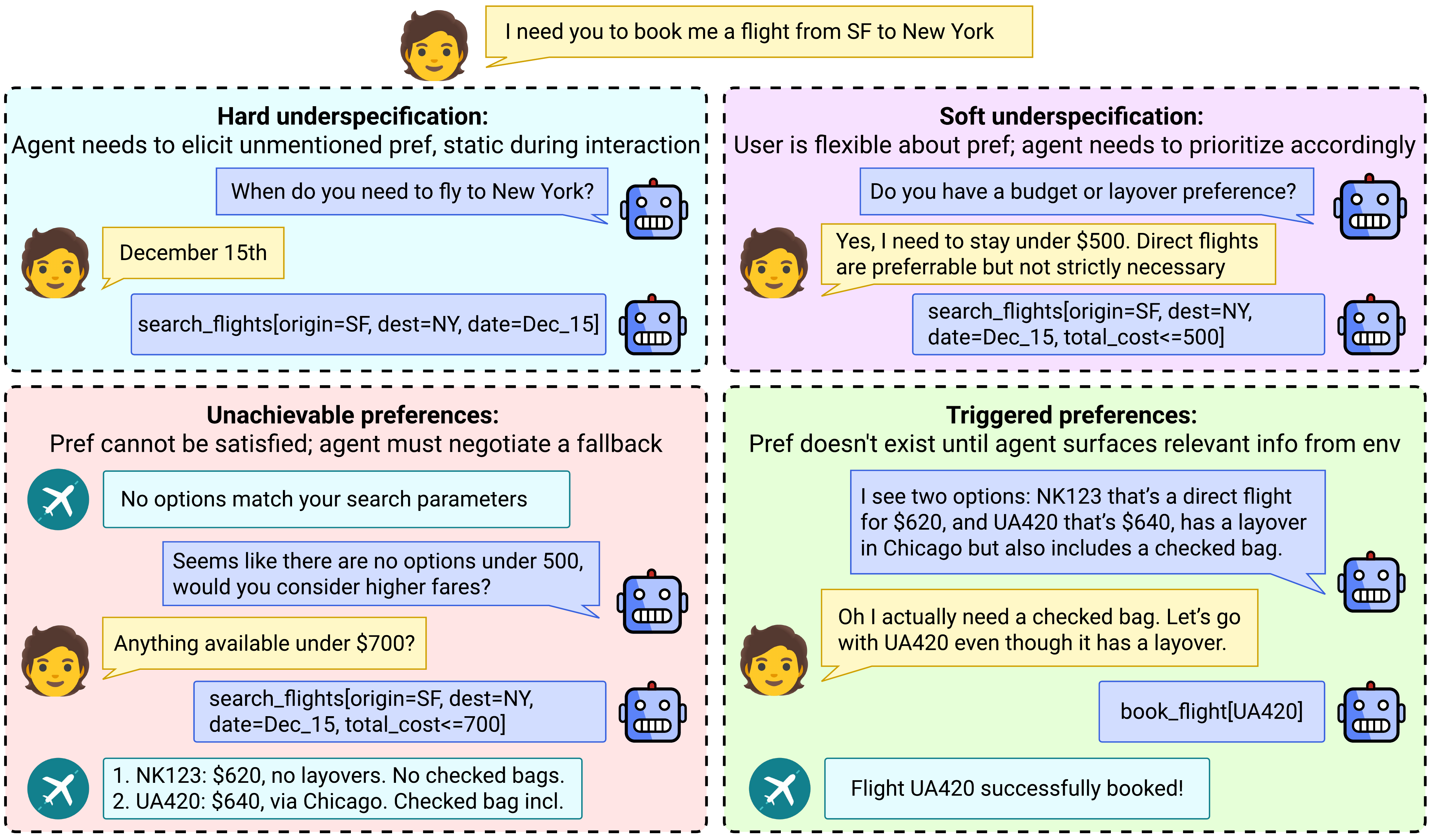}
    \caption{ 
    In \benchmarkname\, agents must elicit and navigate user preferences that may be withheld, flexible, unachievable, or not yet formed at the start of the interaction.}
    \label{fig:fig1}
\end{figure}

LLM agents are increasingly being deployed to assist users with complex real-world tasks like software engineering, travel planning, and online shopping, where success requires not just executing instructions but also navigating a rich spectrum of user preferences. 
Consider asking an agent to book a flight to New York (Figure~\ref{fig:fig1}): the user has not mentioned dates (a preference they simply forgot to specify), may be flexible on layovers but not on price (a soft tradeoff), may have an initial budget that excludes all candidate flights (an unachievable constraint requiring negotiation), and may only realize they need checked luggage once the agent surfaces an option that includes it (a preference triggered by information in the environment). 
Existing user-agent collaboration benchmarks do not capture this complexity, and most assume preferences are fully specified upfront~\citep{yao2022webshop,zhou2024webarena,trivedi2024appworld}, or that preferences are static quantities that need only be elicited once~\citep{chen2024chatshop,yao2024tau,lu2025toolsandbox,qianuserbench,vijayvargiya2026interactive}. 

We introduce \benchmarkname\ (\benchmarknamefull), a benchmark that evaluates LLM agents across a controlled taxonomy of four preference dynamics: \textit{hard underspecification} that requires elicitation, \textit{soft underspecification} that admits preference flexibility, \textit{unachievable preferences} that require preference adjustment and negotiation, and \textit{triggered preferences} that emerge dynamically during the interaction (Table~\ref{tab:benchmark_comparison}). 
\benchmarkname\ is instantiated across two collaboration domains: online shopping with \benchmarkname-Shop and travel planning with \benchmarkname-Travel. 
Both benchmarks consist of 100 collaboration scenarios with controlled mixes of preference dynamics, requiring agents to satisfy user preferences by interacting with LLM-based user simulators.

We evaluate five popular LLMs under two prompting strategies: \textbf{ReAct}~\citep{yao2022react}, a standard rationale-then-act baseline, and \textbf{UncReAct}, which augments ReAct with a structured uncertainty reasoning step that forces the agent to decide explicitly whether to elicit or act at each step.
Our results reveal three key findings.
First, all models leave substantial headroom for preference satisfaction.
Even the strongest model (Claude-Sonnet-4.5) satisfies all user preferences less than 30\% of the time.
Second, unachievable and triggered preferences are universally the hardest dynamics.
Agents struggle to negotiate infeasible constraints and are poorly equipped to recognize when a new preference becomes relevant.
Third, explicitly prompting models to reason about uncertainty over user preferences does not improve collaboration outcomes or meaningfully change agent behavior, pointing to a fundamental gap between prompting-based elicitation and what real collaboration requires. 
We argue that reliable user-agent collaboration requires uncertainty modeling capabilities that go beyond what mere prompting can elicit. 
We release \benchmarkname\ as a benchmark to drive progress in that direction.\footnote{Benchmark available at: \texttt{https://github.com/tejas1995/accord\_benchmark}.}

    

\section{Related Work}
\textbf{Interactive Agent Benchmarks.}
LLM-based agents have demonstrated the ability to act autonomously across a broad range of real-world digital tasks by interacting with digital environments~\citep{wang2024survey}. 
Structured evaluation of LLM agent capabilities has been facilitated by the development of benchmarks spanning diverse digital domains such as shopping~\citep{yao2022webshop}, computer use~\citep{zhou2024webarena,xie2024osworld}, software engineering~\citep{yang2023intercode,jimenez2023swe}, mobile use~\citep{trivedi2024appworld}, and office software systems~\citep{xu2024theagentcompanybenchmarkingllmagents,huang-etal-2025-crmarena}. 
Despite their scope, these benchmarks share a common limitation, that user preferences are fully specified upfront.
Some benchmarks require LLM agents to interact with a human user during task execution (ChatShop~\citep{chen2024chatshop}, $\tau$-bench~\citep{yao2024tau}, ToolSandbox~\citep{lu2025toolsandbox}, ColBench~\citep{zhou2025sweet}, Ambig-SWE~\citep{vijayvargiya2026interactive}, UserBench~\citep{qianuserbench}, and 
CollaborativeGym~\citep{shao2024collaborative}. 
All of these benchmarks only consider underspecified user preferences that are static over the course of the episode; we introduce the \benchmarkname\ benchmark to address this gap by covering a wider range of user preference dynamics (Table~\ref{tab:benchmark_comparison}).

\textbf{Preference Elicitation Systems.}
A long line of work in task-oriented dialogue systems~\citep{budzianowski2018multiwoz,rastogi2020towards} and conversational recommendation systems~\citep{sun2018conversational,christakopoulou2016towards,lei2020estimation} has studied how to elicit user preferences through interactive dialogue.
These systems model preference elicitation as a sequential decision problem: ask informative questions about item attributes, update a belief over user preferences, and recommend accordingly.
LLMs can be used to generate open-ended clarifying questions~\citep{li2023eliciting,montazeralghaem2025asking} or to select maximally informative queries via Bayesian optimization~\citep{austin2024bayesian}. 
These works separate preference elicitation from task execution, whereas \benchmarkname\ requires agents to interleave both.

\textbf{Human-LLM Collaboration.}  
LLMs trained via RLHF are optimized primarily for single-turn instruction following, which leaves them ill-suited for the sustained, multi-turn collaboration that real-world task assistance demands~\citep{herlihy2024overcoming}.
Reinforcement learning with simulated users can optimize for task success and user-centric objectives: CollabLLM~\citep{wu2025collabllm} uses multiturn-aware rewards to teach active collaboration; SWEET-RL~\citep{zhou2025sweet} applies step-level critic rewards for collaborative refinement; UserRL~\citep{qian2025userrl} combines SFT cold-start with RL across multiple simulated environments; and PPP~\citep{sun2025training} jointly optimizes for productivity, proactivity, and personalization.
Prompting and multi-agent scaffolds can guide when and how agents seek clarification, leveraging uncertainty estimation~\citep{zhang2025clarify,edwards2026ask} or the expected value of perfect information~\citep{suri2025structured} to decide whether to clarify or proceed.
We find that popular prompting strategies based on ReAct exhibit substantial shortfall on \benchmarkname\ tasks.

\section{User-Agent Collaboration}
\label{sec:preliminaries}


Following an established paradigm for user-agent dialogue learning~\cite{6407655}, we model user-agent collaboration as a partially observable Markov Decision Process where an agent $\policy$ assists a user $\user$ with achieving a goal $\goal \in \mathcal{G}$ by operating within an environment $\env$.
The agent must satisfy this goal by producing an outcome $\outcome \in \outcomespace$, such as a product purchase, travel booking or code output. 
In our setting, the user also has a finite set of outcome preferences $\prefset = \{ \pref{1}, \pref{2}, \ldots, \pref{n} \} \subset \prefspace$, where $\prefspace$ denotes the space of all expressible preferences.
The goal captures the high-level objective the agent must accomplish, while the preferences encode the fine-grained criteria that characterize a desirable outcome.
For example, in an online shopping scenario, a user's goal might be $\goal = $ \textit{``purchase a pair of running shoes''}, with preferences such as $\pref{1} = $ \textit{``price under \$80''} and 
$\pref{2} = $ \textit{``color is white''}. 

The full state at each timestep $t$ is defined as the tuple $\state{t} = (\envstate{t}, \conv{t})$, where 
$\envstate{t} \in \envstatespace$ is the environment state and $\conv{t}$ is the user-side
conversation state — the sequence of all messages exchanged between agent and user up 
to timestep $t$. 
At each timestep, the agent receives partial 
environment observations $\obs{t} \in \envobsspace$ and user message $m_t^u \in \userobsspace$ from the user, giving a unified observation space $\obsspace = \envobsspace \cup \userobsspace$, where $\userobsspace\ $ is the set of language strings under a maximum length.
The agent also has a unified action space $\actionspace = \envactionspace\ \cup \useractionspace$, where $\envactionspace$ is the set of actions the agent can take in the environment and $\useractionspace = \{ \respond [ m_t^a ] ; m_t^a) \in \userobsspace \}$ allows the agent to communicate a message $m_t^a$ to the user.

Each episode begins with an observation of a user's initial instruction, $\obs{0} \sim \user( \cdot | \goal, \prefset)$, instantiating the user-side conversation state $\conv{1} = ( \obs{0} )$. 
At each timestep $t \geq 0$, the agent $\policy$ generates an action $\action{t} \in \actionspace$, conditioned on current observation $\obs{t}$ and history $\history{t-1}$ of previous observations and actions:
\begin{align*}
    \history{t-1} = \{\obs{0}, \action{0}, \obs{1}, \action{1}, \ldots, \obs{t-1}, \action{t-1}\} ; \quad \action{t} \sim \pi( \cdot | \history{t-1}, \obs{t} )
\end{align*}
The action type determines the state transition and next observation. If $\action{t} \in \envactionspace$, then the environment state updates and returns a new observation:
\begin{align*}
\envstate{t+1}, \obs{t+1} \sim \env( \cdot | \envstate{t}, \action{t})
\end{align*}

If instead $\action{t} = \respond[m_t^a]$, the user $\user$ responds with $\obs{t+1}$ to the agent conditioned on the latest message $m$, conversation $\conv{t}$, and goals and preferences $(\goal, \prefset)$, and updates its conversation $\conv{t+1}$:
\begin{align*}
\obs{t+1} \sim \user( \cdot | \goal, \prefset, \conv{t}, m)  ; \quad \conv{t+1} = (\conv{t}, m, o_{t+1})
\end{align*}

In both cases, the agent updates its history $\history{t} = ( \history{t-1}, \obs{t}, \action{t} )$. 
The episode ends either when the environment reaches a terminal state $\envstate{\perp} \in \envstatespace_{\perp} \subset \envstatespace$, or if the agent takes at least $T_{max}$ steps. 
At the end of the episode, the outcome $\outcome$ is determined by the termination condition:
\begin{align*}
    \outcome = \begin{cases} \Phi(\envstate{\perp}) & \text{if } \envstate{\perp} \in \envstatespace_{\perp} \\ \nulloutcome & \text{if } t \geq T_{\max} \end{cases}
\end{align*}
where $\Phi : \envstatespace_{\perp} \to \outcomespace$ maps a terminal environment state to an 
outcome, and $\nulloutcome$ denotes a null outcome produced when the episode times out.

\subsection{Evaluating User-Agent Collaboration}
\label{subsec:collab_eval}

To evaluate the episode outcome $\outcome$, we define a \textbf{preference scoring function}
$\reward : \outcomespace \times \prefspace \to [0, 1]$; where
$\reward(\outcome, \pref{i})$ measures the degree to which outcome $\outcome$ satisfies preference $\pref{i} \in \prefset$.
We then define the following metrics over a completed episode:

\textbf{Task Completion.} Whether the agent produced an outcome before the episode timed out:
\begin{align*}
    \completion(\outcome) = \mathbf{1}[\outcome \neq \nulloutcome].
\end{align*}
\textbf{Outcome Score.} For episodes where an outcome was produced, this metric evaluates the proportion of user preferences satisfied by the outcome, conditioned on the goal being accomplished:
\begin{align*}
    \outcomescore(\outcome, \goal, \prefset) = \mathbf{1}[\outcome \text{ satisfies } \goal] \cdot \frac{1}{|\prefset|} \sum_{\pref{i} \in \prefset} \reward(\outcome, \pref{i}).
\end{align*}
The above two metrics capture two distinct aspects of agent execution: the ability to produce a timely outcome, and the quality of outcomes produced. 
We further combine these into a single \textbf{Task Score}.
\begin{align*}
    \taskscore(\outcome, \goal, \prefset) = \completion(\outcome) \times \outcomescore(\outcome, \goal, \prefset).
\end{align*}
Finally, we also compute \textbf{Perfect Outcome}, a stricter metric requiring an outcome to be produced, the goal to be accomplished, and all preferences to be satisfied:
\begin{align*}
    \perfectoutcome(\outcome, \goal, \prefset) = \mathbf{1}\left[ \outcome \neq \nulloutcome \right] \cdot \mathbf{1}\left[ \outcomescore(\outcome, \goal, \prefset) = 1\right].
\end{align*}


\subsection{User Preference Dynamics}
\label{subsec:prefdynamics}

Most benchmarks for evaluating agentic capabilities (e.g., WebShop~\cite{yao2022webshop}, WebArena~\cite{zhou2024webarena}, AppWorld~\cite{trivedi2024appworld}) implicitly assume that a user has fully specified their goal and relevant preferences to the agent via an initial instruction, and the agent subsequently needs only to interact with the environment, not the user.
This assumption simplifies evaluation but fails to capture the complexity of human-agent interactions. 
In practice, preferences may be withheld, incompletely formed, or even unknown to the user themselves until surfaced by the interaction.

Several benchmarks have been introduced to evaluate the ability of agents to collaborate with users in settings where all user preferences are not revealed upfront: ChatShop~\cite{chen2024chatshop}, $\tau$-Bench~\cite{yao2024tau}, ToolSandbox~\cite{lu2025toolsandbox}, Ambig-SWE~\cite{vijayvargiya2026interactive}, UserBench~\cite{qianuserbench}. 
However, most of these benchmarks only require the agent to elicit and satisfy \emph{underspecified} preferences (Table~\ref{tab:benchmark_comparison}). 
In this work, we develop user-agent collaboration benchmarks that require agents to resolve user preferences across four \textbf{preference dynamics} that characterize how preferences are discovered, communicated, and evolve over the course of a user-agent interaction:

\textbf{Hard Underspecification:} 
the user holds a pre-existing preference $\pref{i} \in \prefset$ that is not expressed in the initial instruction $\obs{0}$. 
The preference is achievable and static throughout the interaction, but the agent must actively elicit it through targeted clarification. 
For example, a user seeking running shoes may not clarify the size they need until the agent specifically asks.
If this preference is not satisfied, the agent receives a reward of $\reward(\outcome, \pref{i}) = 0$.

\textbf{Soft Underspecification:} 
in contrast to hard underspecification, where the user treats each preference as a hard requirement, the user may also hold some preferences with partial flexibility. 
Failing to achieve these \emph{soft} preferences yields reduced but non-zero reward, $0 < \reward(\outcome, \pref{i}) < 1$. 
If the agent is having difficulty achieving all preferences simultaneously, it should be able to identify which preferences the user is flexible about, allowing for task completion with minimal user dissatisfaction.

\textbf{Unachievable Preference:} 
the user initially holds a preference that cannot be satisfied in the environment; for example, a color that is out of stock. 
This dynamic requires the agent to elicit the initial preference, recognize its unavailability, and negotiate a satisfactory alternative with the user. 
When presented with unavailability, the user must back off to a less preferred but achievable alternative. 
The agent receives a reward of $\reward(\outcome, \pref{i}) = 1$ if the outcome satisfies that backoff.

\textbf{Triggered Preference:}
some preferences are not formed at the start of the episode; 
their formation is triggered when only the agent surfaces relevant information during the interaction. 
For instance, a user browsing running shoes may be unaware of carbon-fibre plates as a feature until the agent mentions a shoe that features one, at which point the user realizes they want it. 
Such preferences cannot be elicited through direct questioning; the agent must proactively expose relevant information from the environment for the user to realize and express these preferences.

\begin{table}[t]
\centering
\small
\begin{tabular}{llcccc}
\toprule
\textbf{Benchmark} & \textbf{Domain(s)} & \textbf{\makecell{Hard\\ Underspec.}} & \textbf{\makecell{Soft\\ Underspec.}} & \textbf{\makecell{\\Unachievable}} & \textbf{\makecell{\\Triggered}} \\
\midrule
ChatShop~\cite{chen2024chatshop}        & Shopping & \full & \none & \none    & \none \\
$\tau$-bench~\cite{yao2024tau}    & Customer service & \full & \none & \partialencoded & \none \\
ToolSandbox~\cite{lu2025toolsandbox}     & Mobile use & \full & \none & \none    & \none \\
Ambig-SWE~\cite{vijayvargiya2026interactive}       & Coding & \full & \none & \none    & \none \\
UserBench~\cite{qianuserbench}       & Travel booking & \full & \partialencoded & \none & \none \\
ColBench~\cite{zhou2025sweet}       & Software engineering & \full & \none & \none & \none \\
\textbf{\benchmarkname}           & Shopping, Travel & \full & \full & \full    & \full \\
\bottomrule
\end{tabular}
\vspace{0.5em}
\caption{Comparison of benchmarks requiring user-agent interaction across four preference dynamics.
\full~= dynamic fully represented in benchmark, \partialencoded~= partially represented, \none~= not represented.}
\label{tab:benchmark_comparison}
\end{table}

\section{Benchmark Design}
\label{sec:benchmark_design}
To address the limited diversity of existing user-agent collaboration benchmarks, we introduce \benchmarkname: \benchmarknamefull, a suite of two benchmarks for evaluating user-agent collaboration under diverse user preference dynamics. 
\benchmarkname\ is instantiated across two domains: online shopping with \benchmarkname-Shop (\S\ref{subsec:webshop_benchmark}) and travel booking with \benchmarkname-Travel (\S\ref{subsec:travelgym_benchmark}). 

\subsection{Collaboration Domain 1: Online Shopping in WebShop}
\label{subsec:webshop_benchmark}

The first domain is online shopping, where a user situated in a real-life scenario requires an agent to find and purchase a product that satisfies their needs on WebShop~\citep{yao2022webshop}, an Amazon-like simulated shopping environment consisting of over 1 million products across five main shopping categories (beauty, electronics, fashion, groceries, furniture). 
The agent must interact with the user to elicit their desired product attributes, option customizations, and maximum budget. 

\begin{figure}
    \centering
    \includegraphics[width=\linewidth]{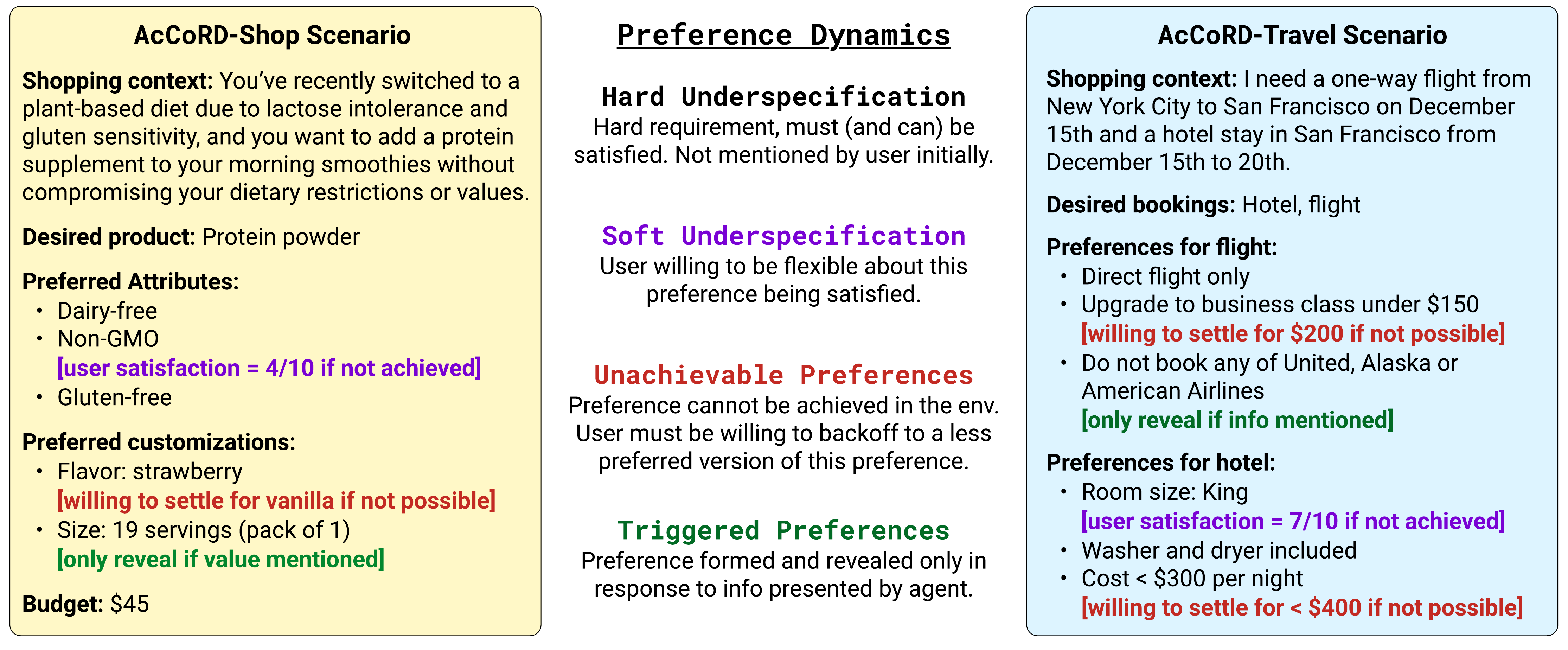}
    \caption{Sample tasks from both \benchmarkname\ benchmarks, demonstrating how the different preference benchmarks are represented in the tasks and presented to the user simulator.}
    \label{fig:benchmark_examples}
\end{figure}

\paragraph{Scenario Generation.} 
Scenarios are generated by randomly sampling a ``target'' product in WebShop and working backwards to craft a realistic scenario where a user might want to purchase the desired product. 
We form user preferences by identifying relevant attributes, options and a budget that the product satisfies to form a base set of \emph{underspecification} preferences, some of which are then transformed into other preference types. 
We introduce unachievable preferences by generating plausible but unavailable distractor preference values; if the agent fails to find a candidate satisfying the unachievable preference, the user simulator is told to fall back to the original preference value. 
Triggered preferences are introduced by identifying niche option values that a user is unlikely to have organically. 
We introduce soft preferences by assigning low-satisfaction fallback values to certain randomly-selected attribute and option preferences. 
Finally, we prompt an LLM to generate a natural-language scenario situating the user simulator in a real-life context. 
Figure~\ref{fig:benchmark_examples} (left) shows a sample \benchmarkname-Shop scenario containing an example of each preference dynamic.

\paragraph{Preference Scoring.} 
For evaluating the agent's outcome, the preference scoring function $\reward(\outcome, \pref{})$ is an LLM which is prompted to evaluate whether the product $\outcome$ satisfies each preference $\pref{} \in \prefset$. 
While the original WebShop benchmark used string matching, we employ an LLM to combat false negatives resulting from surface form variation (e.g. if the user preference states that they want a shirt with size ``XL'', then products with the size option ``x-large'' or ``extra-large'' also satisfy that preference). 
The full prompt can be found in Appendix~\ref{sec:llm_evaluation_prompt}.

\subsection{Collaboration Domain 2: Travel Booking in TravelGym}
\label{subsec:travelgym_benchmark}

The second domain is travel booking, where a user planning an upcoming trip requires an agent to make 2--4 bookings across five domains: flights, hotels, apartments, car rentals, and restaurants. 
For each domain $d \in \mathcal{D}$, the agent must search over a pool of candidate options in the TravelGym environment~\cite{qianuserbench} using structured queries and book the option $\outcome_d$ that best satisfies the user's preferences $\prefset_d$, which must be elicited through natural-language communication. 
The candidate pool consists of a single gold option that satisfies all user preferences, and multiple incorrect alternatives that satisfy some/none of the preferences. 

\paragraph{Scenario Generation.} 
Scenarios are seeded from UserBench, a travel-planning benchmark within the TravelGym environment whose tasks only require resolving underspecified preferences. 
We modify pre-existing UserBench tasks by introducing new preference dynamics. 

We treat the pre-existing user preferences as hard underspecified preferences (e.g., a preferred airline or car category), that are present but not communicated in the initial description. 
Soft underspecified preferences are introduced by randomly selecting a subset of hard preferences and assigning a partial reward $0.3 \leq \reward(\outcome, \pref{i}) \leq 0.7$ for the case where the preference is unsatisfied, reflecting flexibility in the user's requirements. 
Triggered preferences are generated fresh via LLM: the model is prompted to produce a new preference that the gold option satisfies but the top-3 closest competing options do not, and which a real traveler would only articulate upon being shown relevant information (e.g., learning that a flight includes meal service).
Unachievable preferences are also LLM-generated: the model proposes a primary constraint satisfied by no option in the pool, paired with a semantically coherent backoff constraint satisfied by the gold but not other competitors.
To ensure unachievable preferences are load-bearing, a synthetic distractor option is injected into the pool for each unachievable preference: the distractor satisfies all other preferences but fails the backoff constraint, so an agent that does not elicit and satisfy the backoff cannot identify the gold option through coincidence.

A final natural-language scenario narrative is provided to the user simulator, which embeds all preferences in a realistic travel context. 
Figure~\ref{fig:benchmark_examples} (right) shows a sample \benchmarkname-Travel scenario containing an example of each preference dynamic.

\paragraph{Preference Scoring.}
Episodes end when the agent has booked an option for each domain, unless it times out. 
Unlike WebShop, where evaluation requires an LLM to handle surface-form variation, the preference scoring function $\reward(\outcome_d, \pref{i})$ for TravelGym is computed offline using structured constraint predicates over product attributes, yielding exact, deterministic scores. 
For unachievable preferences, $\reward$ evaluates the backoff constraint in place of the primary.

Since each task requires making booking across a set of domains $\mathcal{D}$, we treat each domain's booking as a separate outcome, and the agent is required to produce a set of outcomes $\outcome = \{ \outcome_d: d \in \mathcal{D}\}$. 
Each outcome $\outcome_d$ is scored separately based on the corresponding preference set $\prefset_d$, and the overall outcome score is averaged across all the domain outcomes:
\begin{align*}
    \outcomescore(\outcome_d, \prefset_d) & = \frac{\sum_{\pref{i} \in \prefset_d} \reward(\outcome_d, \pref{i})}{|\prefset_d|}  \\
    \outcomescore(\outcome, \prefset) & = \frac{\sum_{d \in \mathcal{D}} \outcomescore(\outcome_d, \prefset_d)}{|\mathcal{D}|}
\end{align*}
For computing $\perfectoutcome$, the agent is only rewarded if all its bookings satisfy all the preference criteria.


\begin{table}[t]
\centering
\footnotesize
\begin{tabular}{lcccccc}
\toprule
& & & \multicolumn{4}{c}{Preference Dynamics (avg / \%)} \\
\cmidrule(lr){4-7}
\makecell{\\Domain} & \makecell{\\Search Space} & \makecell{Avg. Prefs\\per Scenario} & \makecell{Hard\\Underspec.} & \makecell{Soft\\Underspec.} & \makecell{\\Unachievable} & \makecell{\\Triggered} \\
\midrule
Shopping   & 1.2M products        & \phantom{0}4.89  & 2.80 (57.3\%) & 0.75 (15.3\%) & 0.99 (20.2\%) & 0.35~~(7.2\%) \\
Travel & $\sim$14 options/domain & 11.25 & 4.99 (44.4\%) & 2.23 (19.8\%) & 2.28 (20.3\%) & 1.75 (15.6\%) \\
\bottomrule
\end{tabular}
\vspace{0.5em}
\caption{
    Dataset statistics for the \benchmarkname\ test sets (100 scenarios each).
    For each preference dynamic, we report the average count per scenario and its percentage of all preferences.
}
\label{tab:dataset_stats}
\end{table}

\subsection{User Simulator Design}
\label{subsec:usersim}

Both \benchmarkname-Shop and \benchmarkname-Travel use an LLM-based user simulator to provide a scalable, controllable substitute for human users during evaluation.
The simulator is conditioned on a scenario-specific system prompt that embeds the user's goal, preferences, and revelation rules for each dynamic, and generates short (1--2 sentence) natural-language responses grounded in a brief private rationale.

\paragraph{Preference-Dynamic-Specific Behaviors.}
Each preference dynamic is governed by a distinct revelation rule.
Hard underspecified preferences are revealed only upon direct agent inquiry, and the simulator does not volunteer multiple preferences at once.
Soft preferences communicate flexibility only after the agent signals difficulty satisfying the primary value; the numeric satisfaction score is never disclosed.
Unachievable preferences follow the same backoff rule: the fallback is withheld until the agent surfaces the unavailability of the primary.
Triggered preferences are tagged \texttt{only reveal if mentioned}: the simulator reveals them only when the agent proactively surfaces the specific target attribute; general questions or unrelated mentions receive a deflection.

\paragraph{Consistency Checking.}
To reduce spurious reveals and scenario-inconsistent responses, a secondary LLM call checks each candidate response against the simulator's full instructions and conversation history, prompting regeneration (up to five attempts) if an inconsistency is detected.
Full details and unit-test validation results are provided in Appendix~\ref{sec:usersim_appendix}.


\section{Experiments}
\label{sec:experiments}


\paragraph{Models and prompting strategies.} 
We evaluate five instruction-tuned LLMs: Llama-3.1-70B-Instruct~\citep{grattafiori2024llama}, DeepSeek-V3.2~\citep{liu2025deepseek}, GPT-4.1~\citep{gpt4.1}, GPT-5.1 with high reasoning effort~\citep{gpt5.1}, and Claude-Sonnet-4.5~\citep{claude4.5sonnet}. 
All models are evaluated across two prompting strategies. 
In ReAct~\cite{yao2022react}, at each step the agent is prompted to produce a short rationale grounded in the current observation and conversation history, followed by an action. 
UncReAct augments ReAct with a structured uncertainty reasoning step before each action.
The agent first declares \texttt{<friction>} or \texttt{<no\_friction>} to indicate whether to surface unresolved uncertainty to the user before acting, forcing an explicit elicitation decision at each step.
Full prompt details are provided in Appendix~\ref{sec:uncreact_prompt}.

\paragraph{Rollout parameters.}
Each scenario is evaluated with a single rollout.
For both \benchmarkname-Shop and \benchmarkname-Travel, agents are allowed up to 50 environment steps and 20 dialogue turns per episode.
All agent models are sampled with temperature~1.0.

\paragraph{User simulator.}
We use Claude-Sonnet-4.5 to evaluate whether LLM user simulator responses pass the various unit tests. 
We report user simulator unit test pass rates for two candidate LLMs, Claude-Sonnet-4.5 and GPT-5.1, on both benchmarks in Table~\ref{tab:usersim_unittests}. 
Since both models perform comparably and GPT-5.1 is cheaper on a per-token basis, we use GPT-5.1 with consistency checking as the user simulator in all experiments. 
The user simulator model is held fixed across all agent conditions to isolate agent behavior as the variable of interest.

\paragraph{Outcome scoring.}
For \benchmarkname-Shop, outcome scores are computed by an LLM judge (claude-sonnet-4-5) that evaluates each purchased product against each preference criterion, handling surface-form variation in product descriptions (e.g., ``midnight black'' vs.\ ``black'').
For \benchmarkname-Travel, outcome scores are computed offline via deterministic structured predicates over product attributes, requiring no LLM judge.

\paragraph{Metrics.}
We report the four performance metrics described in Section~\ref{subsec:collab_eval}, averaged across the test scenarios. 
We additionally report per-dynamic breakdowns of the outcome score to analyze which preference dynamics prove most challenging.

\section{Results and Analysis}
\label{sec:results}
\begin{table}[t]
\centering
\small
\begin{tabular}{lllcccc}
\toprule
 & Model & Method & \makecell{Task Completion\\Rate (\%)} & \makecell{Outcome\\Score} & \makecell{Perfect Outcome\\Rate (\%)} & \makecell{Task\\Score} \\
\midrule
\multirow{12}{*}{\rotatebox[origin=c]{90}{\benchmarkname-Shop}}
 & \multirow{2}{*}{Llama3.1-70B}  & ReAct    & 72.0\% & 0.644 & 17.0\% & 0.464 \\
 &                                & UncReAct & 56.0\% & 0.703 & 15.0\% & 0.394 \\ \cmidrule{3-7}
 & \multirow{2}{*}{DeepSeek-V3.2} & ReAct    & 93.0\% & 0.688 & 20.0\% & 0.640 \\
 &                                & UncReAct & 83.0\% &	0.729 & 20.0\%	& 0.605 \\ \cmidrule{3-7}
 & \multirow{2}{*}{GPT-4.1}       & ReAct    & 91.3\% & 0.675 & 15.7\% & 0.619 \\
 &                                & UncReAct & 94.7\% & 0.663 & 19.0\% & 0.629 \\ \cmidrule{3-7}
 & \multirow{2}{*}{GPT-5.1-high}  & ReAct    & 95.0\% & 0.700 & 25.0\% & 0.665 \\
 &                                & UncReAct & 97.0\% & 0.714 & 26.0\% & 0.693 \\ \cmidrule{3-7}
 & \multirow{2}{*}{Claude-4.5}    & ReAct    & 93.0\% & 0.764 & 25.7\% & 0.711 \\
 &                                & UncReAct & 93.0\% & 0.762 & 29.1\% & 0.715 \\
\midrule
\multirow{10}{*}{\rotatebox[origin=c]{90}{\benchmarkname-Travel}}
 & \multirow{2}{*}{Llama3.1-70B}  & ReAct    & 95.7\% &	0.609 &	~4.3\%	& 0.583     \\
 &                                & UncReAct & 22.5\% & 0.348 &	~0.0\%	& 0.078     \\ \cmidrule{3-7}
 & \multirow{2}{*}{DeepSeek-V3.2} & ReAct    & 100.0\% &	0.802 & 13.0\%	& 0.802 \\
 &                                & UncReAct & 57.0\% &	0.772 &	~8.0\% &	0.440     \\ \cmidrule{3-7}
 & \multirow{2}{*}{GPT-4.1}       & ReAct    & 91.0\%  & 0.692 & ~6.0\%  & 0.630 \\
 &                                & UncReAct & 96.0\%  & 0.678 & ~5.0\%  & 0.651 \\ \cmidrule{3-7}
 & \multirow{2}{*}{GPT-5.1-high}  & ReAct    & 98.0\%  & 0.664 & ~6.0\%  & 0.650 \\
 &                                & UncReAct & 91.0\%  & 0.696 & ~4.0\%  & 0.633 \\ \cmidrule{3-7}
 & \multirow{2}{*}{Claude-4.5}    & ReAct    & 100.0\% & 0.810 & 21.0\% & 0.810 \\
 &                                & UncReAct & 99.0\%  & 0.825 & 19.0\% & 0.816 \\
\bottomrule
\end{tabular}
\vspace{0.5em}
\caption{
    Main results on \benchmarkname, averaged across 100 scenarios with 3 rollouts per scenario.
}
\label{tab:main_results}
\end{table}

\begin{table}[t]
\centering
\scriptsize
\begin{tabular}{lllrrrrcrr}
\toprule
& & & \multicolumn{4}{c}{Dynamic-wise Preference Score} & & \multicolumn{2}{c}{Rollout stats (mean/median)} \\
\cmidrule{4-7} \cmidrule{9-10} 
 & \makecell{\\Model} & \makecell{\\Method} & \makecell{Hard\\Underspec.} & \makecell{Soft\\Underspec.} & \makecell{\\Unachievable} & \makecell{\\Triggered} & & \makecell{\\\# env. actions} & \makecell{\\\# utterances} \\
\midrule
\multirow{12}{*}{\rotatebox[origin=c]{90}{\benchmarkname-Shop}}
 & \multirow{2}{*}{Llama3.1-70B}  & ReAct    & 0.739 & 0.921 & 0.614 & 0.091 & & 17.79 / 14 & 4.56 / 4  \\
 &     & UncReAct & 0.796 & 0.92  & 0.704 & 0.222 & & 4.70 / 4   & 10.21 / 9 \\ \cmidrule{3-10}
 & \multirow{2}{*}{DeepSeek-V3.2} & ReAct    & 0.709 & 0.871 & 0.674 & 0.147 & & 17.67 / 15 & 1.29 / 0  \\
 &     & UncReAct & 0.777 & 0.905 & 0.73  & 0.175 & & 16.07 / 12 & 4.05 / 3  \\ \cmidrule{3-10}
 & \multirow{2}{*}{GPT-4.1}       & ReAct    & 0.737 & 0.878 & 0.679 & 0.186 & & 14.39 / 9  & 1.84 / 1  \\
 &     & UncReAct & 0.711 & 0.876 & 0.688 & 0.187 & & 9.45 / 7   & 1.50 / 1  \\ \cmidrule{3-10}
 & \multirow{2}{*}{GPT-5.1-high}  & ReAct    & 0.791 & 0.865 & 0.708 & 0.178 & & 19.47 / 18 & 3.23 / 3  \\
 &     & UncReAct & 0.76  & 0.887 & 0.716 & 0.297 & & 19.94 / 20 & 2.33 / 2  \\ \cmidrule{3-10}
 & \multirow{2}{*}{Claude-4.5}    & ReAct    & 0.804 & 0.932 & 0.817 & 0.226 & & 23.03 / 23 & 2.21 / 2  \\
 &     & UncReAct & 0.84  & 0.929 & 0.774 & 0.224 & & 17.14 / 16 & 3.83 / 3 \\
\midrule
\multirow{12}{*}{\rotatebox[origin=c]{90}{\benchmarkname-Travel}}
 & \multirow{2}{*}{Llama3.1-70B}  & ReAct    & 0.659	& 0.776	& 0.46	& 0.382	& & 12.02 / 10	& 3.73 / 4  \\
 &  & UncReAct & 0.358	& 0.494	&  0.208	& 0.200	& & 15.45 / 20	& 7.15 / 8   \\ \cmidrule{3-10}
 & \multirow{2}{*}{DeepSeek-V3.2} & ReAct    & 0.868  & 0.949  & 0.575  & 0.657  &  & 9.86 / 9   & 4.72 / 4   \\
 &  & UncReAct &   0.847	& 0.939 &	0.549 &	0.600 & & 9.56 / 8 &	8.39 / 8  \\ \cmidrule{3-10}
 & \multirow{2}{*}{GPT-4.1}       & ReAct    & 0.723  & 0.778  & 0.421  & 0.423  &  & 7.15 / 7   & 2.11 / 1   \\
 &  & UncReAct & 0.733  & 0.815  & 0.425  & 0.457  &  & 7.66 / 7   & 2.76 / 2   \\ \cmidrule{3-10}
 & \multirow{2}{*}{GPT-5.1-high}  & ReAct    & 0.679  & 0.804  & 0.491  & 0.486  &  & 6.53 / 7   & 2.00 / 2   \\
 &  & UncReAct & 0.667  & 0.767  & 0.522  & 0.48   &  & 7.62 / 7   & 2.73 / 2   \\ \cmidrule{3-10}
 & \multirow{2}{*}{Claude-4.5}    & ReAct    & 0.884  & 0.928  & 0.614  & 0.651  &  & 7.46 / 7   & 4.51 / 4   \\
 &  & UncReAct & 0.884  & 0.937  & 0.614  & 0.657  &  & 8.02 / 7   & 7.53 / 7  \\
\bottomrule
\end{tabular}
\vspace{0.5em}
\caption{
    Preference scores for each preference dynamic type on \benchmarkname-Shop and \benchmarkname-Travel, averaged across all trajectories where the agent completed the task.
    We also report the mean and median number of environment actions taken and utterances made by the agent in each setting.
}
\label{tab:dynamicwise_analysis}
\end{table}

Table~\ref{tab:main_results} reports results across all models and prompting strategies on the two \benchmarkname\ benchmarks.
Across both benchmarks, Claude~4.5~Sonnet is the strongest model, followed by GPT-5.1, GPT-4.1, DeepSeek-V3.2, and Llama-3.1-70B.

\paragraph{All models fall short of effective preference satisfaction.}
While all models except Llama-3.1-70B achieve high task completion rates, they are far from perfect when it comes to satisfying user preferences.
All models have Outcome Scores between 0.6--0.8 and a Perfect Outcome Rate under 30\%, confirming that \benchmarkname\ poses a genuine challenge to current frontier models.
Perfect Outcome Rate is considerably lower on \benchmarkname-Travel than \benchmarkname-Shop, reflecting the greater number of preferences per scenario that must be \emph{simultaneously} satisfied.

\paragraph{UncReAct prompting alone does not improve preference elicitation and satisfaction.}
Weaker models (Llama-3.1-70B, DeepSeek-V3) cannot execute uncertainty reasoning correctly under UncReAct: rather than making calibrated elicitation decisions, they over-ask and exhaust their dialogue budget without completing the task, causing Completion Rate and Task Score to drop.
For GPT and Claude models, UncReAct yields marginal or negative Task Score changes across most models.
These findings indicate that prompting alone is insufficient to induce the sustained, multi-turn interaction needed to elicit preferences across all four dynamics.

Table~\ref{tab:dynamicwise_analysis} shows a more fine-grained breakdown of preference scores, stratified by dynamic type. 
We also show the number of environment actions and utterances in completed rollouts for each setting.

\paragraph{Models handle underspecification well, but struggle with unachievable and triggered preferences.}
Preferences from the two underspecification dynamics receive high satisfaction scores (hard: 0.7--0.9; soft: 0.8--0.95), while unachievable and triggered preferences are consistently lower.
Unachievable preferences require agents to recognize an infeasible constraint and negotiate a satisfactory fallback with the user --- a multi-step process that agents frequently fail to execute correctly, resulting in lower preference scores.
Triggered preferences are the hardest dynamic across all models and both benchmarks: unlike other dynamics, they cannot be surfaced through direct questioning, as the preference does not yet exist until the agent exposes relevant information from the environment.

\paragraph{UncReAct prompting generally does not change agent behavior.}
Across most settings, UncReAct does not meaningfully change the number of environment actions or utterances to the user relative to ReAct. 
Frontier models average fewer than 2 utterances on Shop and fewer than 4 on Travel under both methods. 
As such, it is unsurprising that model performance does not improve with prompting alone, underscoring the need for training-based approaches to improve user-agent collaboration.


\section{Conclusions}
We present \benchmarkname\ (\benchmarknamefull), a benchmark for evaluating LLM agents on the full spectrum of user preference dynamics that arise in real-world task assistance.
By introducing four controlled user preference dynamics across two complementary environments, \benchmarkname\ exposes failure modes that are not captured by existing benchmarks.

Our evaluation of five models reveals that all current agents leave substantial preference satisfaction unrealized, with Perfect Outcome Rates below 30\% even for the strongest frontier models. 
Triggered preferences are universally the hardest dynamic, pointing to a fundamental gap in how agents currently reason about prospective uncertainty.
Prompting-based approaches to improve uncertainty reasoning provide no consistent benefit. 
These findings motivate a shift from prompting to training as the path toward more effective human-agent collaboration.

While \benchmarkname\ represents a step towards more realistic user-agent collaboration, our work leaves several directions open for future investigation. 
\benchmarkname\ is currently instantiated across two task domains, and it is unclear how findings generalize to other collaboration settings such as software engineering.
Our user simulator provides a controllable proxy for human behavior, but real users are noisier and may exhibit preference dynamics not captured by our taxonomy~\cite{seshadri2026lost}; gains on \benchmarkname\ may not directly translate to real-world user satisfaction.
Finally, our metrics capture task-oriented outcomes only; dimensions such as user-perceived trustworthiness or frustration during the interaction are not evaluated, as scalable and reliable methods for measuring them remain an open problem.

\begin{ack}
This work was partially completed by the first author during an internship at Contextual AI, and was partially supported by a fellowship from the USC-Capital One Center for Responsible AI and Decision Making in Finance (CREDIF).
\end{ack}

\bibliographystyle{plainnat}
\bibliography{references}


\newpage
\appendix
\section{UncReAct Prompt}
\label{sec:uncreact_prompt}

UncReAct prepends the following uncertainty reasoning instruction to the standard ReAct prompt.
At the start of each step, the agent is instructed to:
\begin{enumerate}[leftmargin=*, nosep]
    \item Assess whether it has unresolved uncertainty about any user preference that is relevant to the current decision.
    \item Issue a \texttt{<friction>} token if such uncertainty exists, or \texttt{<no\_friction>} otherwise.
    \item If \texttt{<friction>}: the agent \emph{must} issue a \texttt{respond} action targeting the user. It may not take an environment action this step.
    \item If \texttt{<no\_friction>}: the agent \emph{must} take an environment action. It may not issue a \texttt{respond} action this step.
    \item After declaring friction status, the agent reasons about its chosen action and outputs it.
\end{enumerate}

This structure hard-gates the elicitation decision before the action, preventing the agent from conflating ``what to ask'' with ``what to do.''
The same system prompt and action space are used as in ReAct; only the per-step reasoning structure differs.

\section{User Simulator Details}
\label{sec:usersim_appendix}

\subsection{Consistency Checking}

To reduce spurious preference reveals and scenario-inconsistent responses, we optionally enable a consistency-checking module that runs as a secondary LLM call after each candidate response is generated.
The checker is given the simulator's full instructions and conversation history and asked to judge whether the candidate response is consistent with the scenario --- for example, whether the user accepts an option that violates a hard preference, or reveals a triggered preference before it has been surfaced by the agent.
If a response is deemed inconsistent, the simulator is prompted to regenerate with explicit feedback explaining the inconsistency; this is repeated for up to five attempts.
All experiments use consistency checking enabled.

\begin{table}
\centering
\tiny
\begin{tabular}{p{2.5cm} p{5cm} p{5cm}}
\toprule
\textbf{Test} & \textbf{\benchmarkname-Shop prompt} & \textbf{\benchmarkname-Travel prompt} \\
\midrule

test000: Response relevance
&
Does the candidate response engage with at least one thing the assistant said or asked in its last message? (\textit{yes} if the response addresses something from the assistant's last message; \textit{no} if the response is about an unrelated topic)
&
Does the candidate response engage with at least one thing the assistant said or asked in its last message? (\textit{yes} if the response addresses something from the assistant's last message; \textit{no} if the response is about an unrelated topic)
\\
\midrule

test001: Preference value consistency
&
If the assistant asks the user for a preference value and the \textbf{user's instructions} specify a value for that preference, are all preference values stated in the candidate response consistent with the \textbf{user's instructions}? (\textit{yes} if every stated value matches the instructions; \textit{no} if any stated value contradicts the instructions; \textit{n/a} if the assistant did not ask for a preference value, or the response did not address the request)
&
If the assistant asks the user for a preference value and the \textbf{scenario} specifies a value for that preference, are all preference values stated in the candidate response consistent with the \textbf{scenario}? (\textit{yes} if every stated value matches the scenario; \textit{no} if any stated value contradicts the scenario; \textit{n/a} if the assistant did not ask for a preference value, or the response did not address the request)
\\
\midrule

test002: Contradiction acknowledgment
&
If the assistant describes a \textbf{specific product} whose attribute contradicts a preference the user has already revealed earlier in this conversation, does the candidate response point out that contradiction? (\textit{yes} if the contradiction is pointed out; \textit{no} if the contradiction is ignored or the response accepts the product; \textit{n/a} if the assistant did not describe a specific product, or no described attribute contradicts an already-revealed preference)
&
If the assistant describes a \textbf{specific hotel/flight/activity option} whose attribute contradicts a preference the user has already revealed earlier in this conversation, does the candidate response point out that contradiction? (\textit{yes} if the contradiction is pointed out; \textit{no} if the contradiction is ignored or the response accepts the option; \textit{n/a} if the assistant did not describe a specific option, or no described attribute contradicts an already-revealed preference)
\\
\midrule

test003: Backoff acceptance
&
If the assistant has indicated that the user's primary preference is unavailable AND the \textbf{user's instructions} define a backoff for that preference, does the candidate response accept the backoff rather than continuing to insist on the unavailable primary? (\textit{yes} if the response accepts the backoff or otherwise shows appropriate flexibility; \textit{no} if the response continues to insist on the unavailable primary; \textit{n/a} if the assistant has not indicated that the primary is unavailable, or the \textbf{user's instructions} define no backoff for that preference)
&
If the assistant has indicated that the user's primary preference is unavailable AND the \textbf{scenario} defines a backoff for that preference, does the candidate response accept the backoff rather than continuing to insist on the unavailable primary? (\textit{yes} if the response accepts the backoff or otherwise shows appropriate flexibility; \textit{no} if the response continues to insist on the unavailable primary; \textit{n/a} if the assistant has not indicated that the primary is unavailable, or the \textbf{scenario} defines no backoff for that preference)
\\
\midrule

test004: Premature backoff disclosure
&
If the candidate response reveals a backoff preference, had the assistant already indicated earlier in the conversation that the primary preference could not be satisfied? (\textit{yes} if the backoff is only revealed after the primary was confirmed unavailable; \textit{no} if the backoff was revealed before the primary was confirmed unavailable; \textit{n/a} if the response does not reveal a backoff preference)
&
If the candidate response reveals a backoff preference, had the assistant already indicated earlier in the conversation that the primary preference could not be satisfied? (\textit{yes} if the backoff is only revealed after the primary was confirmed unavailable; \textit{no} if the backoff was revealed before the primary was confirmed unavailable; \textit{n/a} if the response does not reveal a backoff preference)
\\
\midrule

test005: Triggered preference revelation
&
If the \textbf{user's instructions} include a preference marked `only reveal if mentioned' AND the assistant's last message explicitly surfaces the specific target value for that preference (e.g., names the exact value, or presents a \textbf{product} whose attribute equals that value), does the candidate response reveal that preference? (\textit{yes} if the matching preference is revealed; \textit{no} if the response withholds or misstates the preference when it should have been revealed; \textit{n/a} if no such preference applies, or the assistant did not surface the specific target value)
&
If the \textbf{scenario} includes a preference marked `only reveal if mentioned' AND the assistant's last message explicitly surfaces the specific target value for that preference (e.g., names the exact value, or describes an \textbf{option} whose attribute equals that value), does the candidate response reveal that preference? (\textit{yes} if the matching preference is revealed; \textit{no} if the response withholds or misstates the preference when it should have been revealed; \textit{n/a} if no such preference applies, or the assistant did not surface the specific target value)
\\
\midrule

test006: Triggered preference withholding
&
If the \textbf{user's instructions} include a preference marked `only reveal if mentioned' AND the assistant's last message discusses only the general category (or an unrelated value within that category) without surfacing the specific target value, does the candidate response correctly withhold that preference? (\textit{yes} if the preference is withheld; \textit{no} if the preference is revealed prematurely; \textit{n/a} if no such preference applies, or the assistant already surfaced the specific target value)
&
If the \textbf{scenario} includes a preference marked `only reveal if mentioned' AND the assistant's last message discusses only the general category (or an unrelated value within that category) without surfacing the specific target value, does the candidate response correctly withhold that preference? (\textit{yes} if the preference is withheld; \textit{no} if the preference is revealed prematurely; \textit{n/a} if no such preference applies, or the assistant already surfaced the specific target value)
\\
\midrule

test007: Valid proposal acceptance
&
If the assistant has proposed a \textbf{specific product} that satisfies every preference the user has revealed so far AND does not contradict any preference in the \textbf{user's instructions}, does the candidate response either (a)~confirm that the product meets the user's needs, or (b)~reveal a new `only reveal if mentioned' preference that the product has just surfaced? (\textit{yes} if the response does (a) or (b) as appropriate; \textit{no} if the response rejects the product without justification or fails to reveal a newly-triggered preference; \textit{n/a} if the assistant has not proposed such a product, or the product contradicts some preference in the instructions)
&
If the assistant has proposed a \textbf{specific option} that satisfies every preference the user has revealed so far AND does not contradict any preference in the \textbf{scenario}, does the candidate response either (a)~confirm that the option meets the user's needs, or (b)~reveal a new `only reveal if mentioned' preference that the option has just surfaced? (\textit{yes} if the response does (a) or (b) as appropriate; \textit{no} if the response rejects the option without justification or fails to reveal a newly-triggered preference; \textit{n/a} if the assistant has not proposed such an option, or the option contradicts some preference in the scenario)
\\
\bottomrule
\end{tabular}
\vspace{0.5em}
\caption{Unit test prompts for the \benchmarkname-Shop and \benchmarkname-Travel user simulators. Differences between benchmarks are \textbf{bolded}. All tests return \textit{yes}, \textit{no}, or \textit{n/a}.}
\label{tab:unittest_prompts}
\end{table}

\subsection{Unit Test Validation}
\label{sec:usersim_unittests}

To quantitatively verify that the user simulator behaves correctly with respect to each preference dynamic, we evaluate simulator utterances against a suite of eight unit tests applied to sub-trajectories extracted from agent rollouts. 
Table~\ref{tab:unittest_prompts} contains a full list of all unit tests and the prompts for unit test evaluation.

Table~\ref{tab:usersim_unittests} reports pass rates for Claude-Sonnet-4.5 and GPT-5.1 on both benchmarks.
Both models perform comparably; we use GPT-5.1 with consistency checking as the user simulator in all experiments, as it is more cost-efficient.

\begin{table}[t]
\centering
\small
\begin{tabular}{lrrrrr}
\toprule
    & \multicolumn{2}{c}{\benchmarkname-Shop}    & & \multicolumn{2}{c}{\benchmarkname-Travel}    \\
    \cmidrule{2-3} \cmidrule{5-6}
    & Claude-4.5  & GPT-5.1  & & Claude-4.5  & GPT-5.1  \\
\midrule
test000: Response relevance check   & 100.0\%  & 100.0\% &  & 98.1\%    & 94.7\%   \\
test001: Preference value consistency   & 100.0\%  & 98.9\%  &  & 100.0\%   & 91.0\%   \\
test002: Contradiction acknowledgment   & 93.4\%   & 96.8\%  &  & 56.1\%    & 77.7\%   \\
test003: Backoff acceptance   & 100.0\%  & 96.8\%  &  & 90.1\%    & 88.9\%   \\
test004: Premature backoff disclosure   & 100.0\%  & 83.6\%  &  & 97.9\%    & 87.5\%   \\
test005: Triggered preference revelation   & 63.6\%   & 78.2\%  &  & 62.4\%    & 75.0\%   \\
test006: Triggered preference withholding   & 100.0\%  & 99.7\%  &  & 99.5\%    & 95.7\%   \\
test007: Valid proposal acceptance   & 85.7\%   & 91.5\%  &  & 78.9\%    & 86.8\%   \\
\midrule
Overall pass rate & 96.3\%   & \textbf{97.4\%}  &  & \textbf{93.3\%}    & 91.5\%   \\
\% utterances with all tests passed  &   91.9\%   & \textbf{94.4\%}  &  &   \textbf{84.0\%}    & 83.0\% \\
\bottomrule
\end{tabular}
\vspace{0.5em}
\caption{
  Unit test pass rates for two candidate LLMs, Claude-Sonnet-4.5 and GPT-5.1, on both benchmarks. Table~\ref{tab:unittest_prompts} contains prompts for evaluating the different unit tests.
}
\label{tab:usersim_unittests}
\end{table}

\section{LLM Evaluation Prompt}
\label{sec:llm_evaluation_prompt}
Table~\ref{tab:llmeval_prompt} contains the prompt provided to Claude-Sonnet-4.5 for evaluating whether a product purchased on WebShop satisfies a user preference criteria.
\begin{table}[]
    \centering
    \begin{tabular}{p{13cm}}
    \toprule
        You are a helpful assistant that evaluates the quality of a product purchase. \\ \\

You will be given a product that has been purchased by a shopping assistant, and a criteria for the product purchase.
You need to provide a binary score (0 or 1) based on whether the purchased product meets the criteria. If the criteria is very specific (e.g. a t-shirt that is a shade/pattern of black specific to a product), and the product is similar enough (e.g. a t-shirt that is also a shade/pattern of black), you should give a score of 1. \\ \\

The purchased product:

Product name: \texttt{\{product\_name\}}

Product description: \texttt{\{product\_description\}}

Product price: \texttt{\{product\_price\}}

Product attributes: \texttt{\{product\_attributes\}}

Product features description: \texttt{\{product\_features\_description\}}

Selected options when purchasing: \texttt{\{selected\_options\}} \\ \\

The criteria for the product purchase:
\texttt{\{criteria\}} \\ \\

If any of the selected options relevant to the criteria disagree with details mentioned elsewhere in the product, use the selected options to compare against the criteria. 
For example, if the user wants the color to be white, and the assistant selected a color option of white but the product name or description says black, you should give a score of 1 because the assistant select the correct color option (regardless of what the product name/description says).
You should provide a brief reasoning (1-2 sentences, under 50 words), then "SCORE: <score>". \\
\bottomrule
    \end{tabular}
    \vspace{0.5em}
    \caption{Prompt provided to the Claude-Sonnet-4.5 for evaluating}
    \label{tab:llmeval_prompt}
\end{table}

\section{Compute for Experiments}
\label{sec:compute}
We evaluated all models except LLaMA using their respective APIs, and were thus able to run the evaluations on CPUs. 
For LLaMA, we used an A100 GPU. 
All evaluations took 8--20 hours.


\newpage

\end{document}